\documentclass[journal]{IEEEtran}

\usepackage{graphicx,epsfig,subfigure,epstopdf}  
\usepackage{amsfonts,amssymb,amsmath,latexsym,amsthm}
\usepackage{bm}
\usepackage[ruled,linesnumbered]{algorithm2e}
\usepackage{algpseudocode}
\usepackage{multirow}
\usepackage{array}
\usepackage{float}
\usepackage{color}
\usepackage{url}
\usepackage{hyperref}
\usepackage[square,comma,numbers,sort&compress]{natbib}
\usepackage{booktabs}
\usepackage{enumitem} 

\DeclareMathOperator*{\argmax}{arg\,max}

\begin{document}
\title{A Dirichlet Process Mixture of Robust Task Models for Scalable Lifelong Reinforcement Learning}

\author{Zhi~Wang,~\IEEEmembership{Member,~IEEE},~Chunlin~Chen,~\IEEEmembership{Senior Member,~IEEE},~Daoyi~Dong,~\IEEEmembership{Senior Member,~IEEE}
	\thanks{Manuscript accepted by \textit{IEEE Transactions on Cybernetics}, 2022, DOI: DOI: 10.1109/TCYB.2022.3170485.}
\thanks{This work was supported in part by the National Natural Science Foundation of China under Grant 62006111 and Grant 62073160; in part by the Australian Research Council's Discovery Projects funding scheme under Project DP190101566; and in part by the Natural Science Foundation of Jiangsu Province of China under Grant BK20200330. (\textit{Corresponding author: Chunlin Chen.})}
\thanks{Z. Wang and C. Chen are with the School of Management and Engineering, Nanjing University, Nanjing 210093, China (email: \{zhiwang, clchen\}@nju.edu.cn).}
\thanks{D. Dong is with the School of Engineering and Information Technology, University of New South Wales, Canberra, ACT 2600, Australia (email: daoyidong@gmail.com).}
}

\maketitle

\begin{abstract}
While reinforcement learning (RL) algorithms are achieving state-of-the-art performance in various challenging tasks, they can easily encounter catastrophic forgetting or interference when faced with lifelong streaming information. In the paper, we propose a scalable lifelong RL method that dynamically expands the network capacity to accommodate new knowledge while preventing past memories from being perturbed. We use a Dirichlet process mixture to model the non-stationary task distribution, which captures task relatedness by estimating the likelihood of task-to-cluster assignments and clusters the task models in a latent space. We formulate the prior distribution of the mixture as a Chinese restaurant process (CRP) that instantiates new mixture components as needed. The update and expansion of the mixture are governed by the Bayesian non-parametric framework with an expectation maximization (EM) procedure, which dynamically adapts the model complexity without explicit task boundaries or heuristics. Moreover, we use the domain randomization technique to train robust prior parameters for the initialization of each task model in the mixture, thus the resulting model can better generalize and adapt to unseen tasks. With extensive experiments conducted on robot navigation and locomotion domains, we show that our method successfully facilitates scalable lifelong RL and outperforms relevant existing methods.
\end{abstract}

\begin{IEEEkeywords}
Chinese restaurant process, Dirichlet process mixture, domain randomization, expectation maximization, lifelong reinforcement learning.
\end{IEEEkeywords}

\section{Introduction}
\IEEEPARstart{L}{ifelong} learning, also referred to as continual learning, corresponds to the capability of continually accommodating new information throughout the lifespan without forgetting previous knowledge, which is crucial for artificially intelligent agents performing in real-world scenarios and proceeding multiple tasks in sequence~\citep{parisi2019continual}
An effective lifelong learning system must satisfy two potentially conflicting goals of stably maintaining old skills and rapidly acquiring a new skill.
These simultaneous constraints represent the long-standing challenge as \textit{stability-plasticity dilemma}~\citep{polikar2001learn}.
While reinforcement learning (RL) algorithms~\citep{sutton2018reinforcement} have achieved state-of-the-art performance in various challenging tasks~\citep{vinyals2019grandmaster,xu2018reinforcement,zhang2018data,wang2019tmechl,wang2019reinforcement,tan2019cooperative,li2020quantum}, they typically exhibit poor sample efficiency and generalization ability when trained on sequential tasks, since continual acquiring new information from a non-stationary task distribution can easily result in catastrophic forgetting or interference~\citep{rolnick2019experience}.
Learning systems are trained to keep outputs consistent with inputs using explicit or implicit parametric function approximation.
Training them towards a new objective will change the data distribution and lead to abrupt erasure of previously acquired knowledge, resulting in high plasticity but little stability. 

Previous attempts to alleviate catastrophic forgetting in lifelong RL settings can generally be classified into three categories: replay-based~\citep{shin2017continual,rolnick2019experience}, regularization-based~\citep{kirkpatrick2017overcoming,kaplanis2019policy}, and expansion-based~\citep{rusu2016progressive,glatt2020decaf}.
Replay-based approaches use a replay buffer to store old samples, which are reproduced for rehearsal and interleaving online updates when learning a new task.
They require large working memory to store and replay old samples, which might not be viable in real-world situations~\citep{schwarz2018progress}. 
Regularization-based approaches retain old knowledge by adding regularization terms that impose constraints on the update of network weights and prevent large changes on significant weights.
With a limited amount of neural resources, comprising additional loss terms can result in a trade-off on the accomplishment of the old and new tasks~\citep{parisi2019continual}.
In contrast, expansion-based approaches differ from the others in that they dynamically expand the model architecture, e.g., a policy/option library~\citep{glatt2017policy,brunskill2014pac} or the network capacity~\citep{yoon2018lifelong}, upon the arrival of each task to accommodate new knowledge.
Therefore, they can mitigate catastrophic forgetting by avoiding the perturbation on past memories from the new information~\citep{lee2020neural}.
However, previous expansion-based approaches typically suffer from the lack of scalability due to two critical limitations: (i) they heavily rely on explicit task boundaries and hand-designed heuristics for incorporating new resources; (ii) the network size may scale quadratically in the number of encountered tasks~\citep{rusu2016progressive}.

Humans can continually accommodate new information and expand cognitive capabilities while preventing past memories from being perturbed.
For the purpose of artificial general intelligence (AGI), RL algorithms ought to continually build on their experiences to develop increasingly complex skills and adapt quickly to new tasks throughout their lifetime~\citep{koga2015stochastic}, without forgetting what has already been learned.
In the paper, we aim at a novel expansion-based method for scalable lifelong RL, with the assumption that task boundaries are not provided explicitly.
We develop and maintain a Dirichlet process mixture of task models to tackle the non-stationary task distribution, which captures task relatedness by estimating the likelihood of assigning each task to mixture components and clusters the task models in a latent space.
We formulate the prior distribution of the mixture as a Chinese restaurant process (CRP) that assigns some probability of instantiating a new task model as needed.
During lifelong learning, the mixture model is updated via an expectation maximization (EM) procedure, where the E-step calculates the posterior inference of task-to-cluster probabilities and the M-step updates all model parameters for future learning.
Furthermore, we adopt the domain randomization technique to train \textit{robust} prior parameters for the initialization of each task model in the mixture, thus the resulting model can better generalize and adapt to unseen tasks.

Our primary contribution is a scalable lifelong RL method that uses an EM procedure to learn a Dirichlet process mixture of robust task models with a flexible memory system, where the prior distribution of the mixture is formulated as a CRP.
With explicitly estimating task relatedness, our method has the potential to enhance the stability of past memories by modulating transferability across similar tasks, and to promote plasticity by recognizing outlier tasks that require a more significant degree of adaptation. 
The mixture is updated and expanded under the Bayesian non-parametric framework that dynamically adapts the model complexity over the agent's lifetime, instead of fixing the model complexity beforehand in parametric approaches.
By treating the task-to-cluster assignments as latent variables, our method is capable of adapting to the non-stationary task distribution without task boundaries or hand-designed heuristics for incorporating new resources.
Our method is evaluated in the context of deep deterministic policy gradient (DDPG) algorithm on robot navigation and MuJoCo~\citep{todorov2012mujoco} locomotion domains in lifelong learning settings.
Experimental results verify that our method facilitates efficient lifelong RL and outperforms several baseline methods.

The remainder of the paper is structured as below. 
Section~\ref{preliminaries} introduces preliminaries of RL algorithms.
Section~\ref{method} presents the problem statement and our method in detail.
Section~\ref{experiments} shows the experimentation on the robot navigation and MuJoCo locomotion domains.
Section~\ref{related} reviews related work regarding lifelong RL, and Section~\ref{conclusions} presents concluding remarks and future work.

\section{Preliminaries}\label{preliminaries}
\subsection{Reinforcement Learning (RL)}
The standard paradigm of an RL agent interacting with an environment is formalized as a Markov decision process (MDP) $\langle \mathcal{S}, \mathcal{A}, \mathcal{T}, \mathcal{R}, \gamma\rangle$, where $\mathcal{S}$ and $\mathcal{A}$ denote the state and action spaces, respectively, $\mathcal{T}:\mathcal{S}\times\mathcal{A}\times\mathcal{S}\to[0,\infty)$ defines the probability density function of transitioning to state $s_{t+1}\in\mathcal{S}$ conditioned on the agent taking action $a_t\in\mathcal{A}$ in state $s_t\in\mathcal{S}$, $\mathcal{R}: \mathcal{S}\times\mathcal{A}\to\mathbb{R}$ is the reward function that maps each transition $(s_t,a_t)$ to a scalar, and $\gamma\in[0,1)$ is the discounting factor.
RL aims to learn a policy, a probability density function over available actions given a state $\pi(a_t|s_t)$, that maximizes the expected return as
\begin{equation}
\pi^* = \argmax_{\pi}\mathbb{E}_{\pi}\left[\sum_{t=0}^{\infty}\gamma^tr_t\right],
\end{equation}
where $r_t\sim r(s_t, a_t)$ denotes the received reward after taking action $a_t$ in state $s_t$.

Model-free methods directly interact with an initially unknown environment to learn optimal policies, releasing the dependency on an explicit model or any prior knowledge of the environment.
Off-policy methods decouple the behavior and target policies, enabling an agent to learn using samples collected by arbitrary policies or from replay buffers.
In our method, we utilize the deep deterministic policy gradient (DDPG)~\citep{lillicrap2016continuous} algorithm, a popular variant of the model-free off-policy Q-learning~\citep{watkins1992q} algorithm for continuous control.

\subsection{Q-learning}
The expected return is related to the optimal action-value function as
\begin{equation}
Q^*(s,a) = \max_{\pi}\mathbb{E}_{\pi}\left[\sum_{t'=t}^{\infty}\gamma^{t'-t}r_{t'}\vert s_t=s,a_t=a\right],
\end{equation}
which is the maximal sum of rewards $r_{t'}\sim r(s_{t'}, a_{t'})$ multiplied by the discount factor $\gamma$ at each step $t'$, after executing action $a$ in state $s$.
The optimal Q-function obeys a significant identity, i.e., the Bellman equation~\citep{bellman1966dynamic}:
\begin{equation}
Q^*(s,a)=\mathbb{E}_{s'}\left[r+\gamma\max_{a'}Q^*(s',a')|s,a\right],
\end{equation}
where $r\sim r(s,a)$.
In tabular cases, the widely used Q-learning algorithm~\citep{watkins1992q} updates the action-value function as
\begin{equation}
Q(s,a)\leftarrow Q(s,a) + \alpha (r+\gamma\max_{a'}Q(s',a')-Q(s,a)),
\end{equation}
where $\alpha\in(0,1]$ is the learning rate.

\subsection{Deep Deterministic Policy Gradient (DDPG)}
For generalization in continuous state spaces, we usually use a function approximator to approximate the action-value function, $Q_{\bm{\varphi}}(s,a)\approx Q^*(s,a)$, where $\bm{\varphi}$ denote the learning parameters.
In deep RL, a deep neural network (DNN) is utilized to approximate the Q-function, known as the famous deep Q-network (DQN)~\citep{mnih2015human}.
The parameters $\bm{\varphi}$ are adjusted at each iteration to minimize the mean-squared error (MSE) in the Bellman equation, i.e., Bellman residual.
This induces a loss function $\mathcal{L}(\bm{\varphi})$ that varies at every learning iteration:
\begin{equation}
\mathcal{L}(\bm{\varphi})=\mathbb{E}_{s,a,r,s'}\left[\left(r+\gamma\max_{a'}Q_{\bm{\varphi}}(s',a')-Q_{\bm{\varphi}}(s,a)\right)^2\right].
\end{equation}


To be applicable to continuous action spaces, DDPG~\citep{lillicrap2016continuous} learns a deterministic neural network policy $\mu_{\bm{\phi}}$ (i.e., the actor) along with the action-value function $Q_{\bm{\varphi}}$ (i.e., the critic) by performing gradient updates on parameter sets $\bm{\varphi}$ and $\bm{\phi}$.
Analogous to the classical Q-learning, the critic is trained to minimize the Bellman residual over all sampled transitions as
\begin{equation}
\mathcal{L}(\bm{\varphi}) = \mathbb{E}_{s,a,r,s'}[(r+\gamma Q_{\bm{\varphi}}(s', \mu_{\bm{\phi}}(s'))-Q_{\bm{\varphi}}(s,a))^2].
\end{equation}
The actor is then trained to yield actions that maximize the Q-values estimated by the critic, equivalent to minimizing the loss function as
\begin{equation}
\mathcal{L}(\bm{\phi}) = -\mathbb{E}_{s,a,r,s'}[Q_{\bm{\varphi}}(s, \mu_{\bm{\phi}}(s))].
\end{equation}
While DDPG trains a deterministic policy, its behavior policy used to collect transitions during training is usually augmented with a Gaussian (or Ornstein-Uhlenbeck) noise.
Therefore, actions are collected as $a\sim \mathcal{N}(\mu_{\bm{\phi}}(s), \varsigma^2)$ for fixed standard deviation $\varsigma$.

\section{Method}\label{method}
In this section, we first formulate the problem statement of lifelong RL.
Then, we explain the idea of modeling the latent task structure with a mixture model to deal with the non-stationary task distribution.
Next, we present the non-parametric Bayesian inference framework that formulates the prior distribution over the mixture of task clusters as a CRP and updates the mixture using the EM algorithm.
Finally, we introduce the domain randomization approach that trains the robust prior parameters for each task model.

\subsection{Problem Statement}
Let $\bm{\varphi}$ and $\bm{\phi}$ denote the weights of the deep Q-network (critic) and the policy network (actor), respectively, and $\bm{\theta}=(\bm{\varphi}, \bm{\phi})$.
The model receives the state-action pair $(s,a)$ as its input $x$ and produces an action-value function prediction $Q_{\bm{\varphi}}(s,a)$ as its predicted output $\hat{y}$.
The target value $r+\gamma Q_{\bm{\varphi}}(s',\mu_{\bm{\phi}}(s'))$ can be considered as the ground truth label $y$ to mimic a supervised learning setting, where $r$ and $s'$ are the received reward and the next state when taking action $a$ in state $s$.
The lifelong learning scenario deals with an infinite sequence of tasks $\mathcal{D} = [\mathcal{D}_1, \mathcal{D}_2,...]$ where each task $\mathcal{D}_t$ is associated with a batch of transitions $\mathcal{T}_t=\sum_i(s_i,a_i,r_i,s_i')$.
The tasks change over the lifetime, leading to a non-stationary task distribution, and the current task identity at each time period $t$ is unknown to the learner.
The learner has to perform all tasks in the sequence.
The full objective is thereby given as to minimize the unbiased sum of losses among all tasks as
\begin{equation}
\begin{aligned}
&\mathcal{L}(\bm{\theta}) = \mathcal{L}(\bm{\varphi}, \bm{\phi}) =\mathbb{E}_{\mathcal{D}_t\sim\mathcal{D}}\left[\mathbb{E}_{x,y\sim\mathcal{D}_t}\left[(\hat{y}- y)^2\right]\right]=\\
&\mathbb{E}_{\mathcal{D}_t\sim\mathcal{D}}\left[\mathbb{E}_{s,a,r,s'\sim\mathcal{D}_t}\left[(r+\gamma Q_{\bm{\varphi}}(s',\mu_{\bm{\phi}}(s'))-Q_{\bm{\varphi}}(s,a))^2\right]\right],
\end{aligned}
\end{equation}
which is equivalent to minimizing the Bellman residual over the given transitions.
While being trained for the task at time period $t$, the learner is fed with samples only from task $\mathcal{D}_t$.

In real-world applications, tasks might correspond to customized requirements, user preferences, unknown dynamics of the system, or other unexpected perturbations.
This problem formulation involves a wide variety of RL challenges requiring lifelong adaptation to sequential tasks and balance between plasticity and stability.
Throughout the lifetime, the learner needs to continually build upon previously learned knowledge to facilitate optimizing the policy of the current task at hand, in conjunction with accommodating the acquired new information for future learning.

\subsection{Modeling Latent Task Structure}
Changing circumstances and unpredictable perturbations are quite common in real-world scenarios, resulting in heterogeneous task distributions.
Assuming a single model for lifelong RL is not suitable because it is unlikely to adequately adapt the learner to various tasks using only a few gradient steps.
Expansion-based lifelong learning approaches follow the idea that if we learn a new task with new parameters and keep previous parameters unchanged, we can well preserve the knowledge of previous tasks.
A straightforward way is to train and store separate parameters for each task~\citep{rusu2016progressive}, while it is rather restricted to ideal settings with explicit task boundaries. 
Moreover, it quickly suffers from the lack of scalability as the number of encountered tasks increases.
In contrast, capturing task relatedness is promising to enhance the stability of past memories by modulating both positive and negative transfer, and to promote plasticity by recognizing outlier tasks that require a more significant degree of adaptation~\citep{xue2007multi,jerfel2019reconciling}.
Therefore, we begin with a more rational idea that clusters previous tasks into a mixture of Bayesian models with an appropriate notion of task relatedness, reducing redundancy within past memories.

For handling the task variability, we assume that the parameters of each task model are drawn from a Dirichlet process mixture of base distributions $\{\bm{\theta}^{(l)}\}_{l=1}^{L}$, where $\bm{\theta}^{(l)}$ denote the model parameters corresponding to the $l$-th cluster.
Then, we estimate the relatedness across tasks by calculating the likelihood of task-to-cluster assignments, which equals to clustering the task models in a latent space.
Each task cluster is initialized by some prior parameters $\bm{\theta}^-$, which is learned by the domain randomization technique and will be introduced in more detail in Section~\ref{dr}.
The prior distribution $P(\bm{\theta})$ over the mixture of task models is formulated as a CRP that allows for some probability of instantiating a new cluster as needed. 
Without knowing the number of task clusters, we start with a single mixture component and initialize this task model from $\bm{\theta}^-$.
From here, we continually maintain and update the mixture model to handle the lifelong task distribution, and instantiate new mixture components as required using the CRP.

\begin{figure}[tb]
	\centering
	\includegraphics[width=0.98\columnwidth]{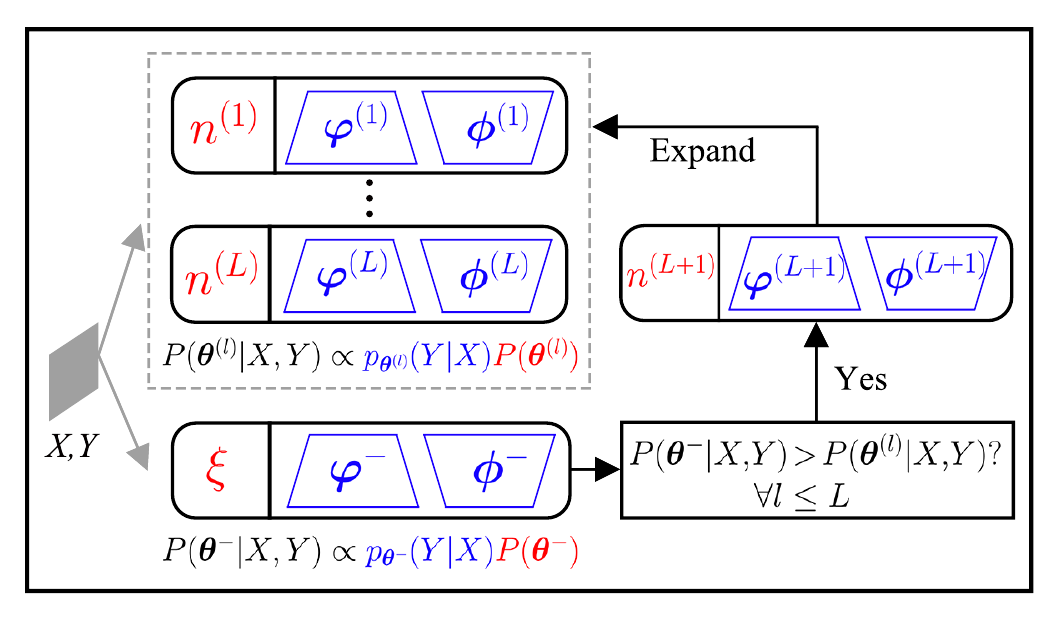}
	\caption{Overview of our method. $n^{(l)}$ is the assigned task count per cluster, and $\xi$ is a constant that regulates the instantiation of new clusters.}
	\label{fig:method}
\end{figure}

Existing approaches usually rely on awareness of explicit task identities, which is unlikely to hold in real-world applications.
Instead, we use the mixture model to estimate the prior and posterior distributions over task clusters, which are utilized to predict task identifies and update parameters of all task models.
This results in a scalable lifelong RL method that jointly learns task-to-cluster assignments and model parameters, which can efficiently modulate the task transferability by clustering task models in a latent space.
The main idea of our method is illustrated as in Fig.~\ref{fig:method}.

\subsection{A Dirichlet Process Mixture for Modeling Task Distribution}
Let the categorical latent variable $z$ denote the cluster assignment of task model parameters $\bm{\theta}$.
Since $z$ is unknown, we ought to infer the posterior task-to-cluster assignments $P(z|X, Y)$, where $(X,Y)=\sum_i(x_i,y_i)$ is the dataset constructed from a batch of transitions at a given time period.
Moreover, we cannot know the total amount of task clusters in advance.
Hence, we propose to employ a Bayesian non-parametric framework, specifically the Dirichlet process mixture model (DPMM), to formulate the non-stationary task distribution into a flexible structure where task clusters are dynamically established and expanded throughout lifelong learning.
The instantiation of DPMM is depicted by the Chinese restaurant process (CRP) that is well suitable for lifelong learning.
The CRP is a discrete-time stochastic process analogous to seating an infinite sequence of customers at tables in a Chinese restaurant, where each table represents a distinct cluster. 
Each customer chooses a pre-existing table with a probability proportional to the count of customers already seated there, or sits down alone at a new empty table with a probability proportional to a preset concentration parameter.

For an infinite sequence of tasks $\mathcal{D}=[\mathcal{D}_1,\mathcal{D}_2,...]$, the first task is allocated to the nominal cluster since the number of task clusters is unknown.
At time period $t$, assume that the accumulated knowledge from previous time periods $1,2,...,t-1$ is accommodated as a mixture of $L$ task clusters $\{\bm{\theta}_t^{(l)}\}_{l=1}^{L}$.
Then, the prior distribution of cluster assignments for the current task is given by
\begin{equation}
P(\bm{\theta}_t^{(l)}) = P(z_t=l) = \left\{\begin{matrix}
\frac{n^{(l)}}{t-1+\xi}, & l\le L \\
\frac{\xi}{t-1+\xi}, & l=L+1,
\end{matrix}\right.
\end{equation}
where $n^{(l)}$ is the expected number of tackled tasks that have occupied the $l$-th cluster, and $\xi$ is a constant positive concentration parameter for regulating the new cluster instantiation.
$l\le L$ indicates assigning the current task to an existing cluster, and $l=L+1$ implies the potential spawning of a new task cluster into the mixture model.
Taking all history periods into account, the prior probabilities over task clusters become
\begin{equation}
P(\bm{\theta}_t^{(l)}|\bm{\theta}_{1:t-1}, \xi)=\left\{\begin{matrix}
\frac{\sum_{t'=1}^{t-1}P(\bm{\theta}_{t'}^{(l)})}{t-1+\xi}, & l\le L \\
\frac{\xi}{t-1+\xi}, & l=L+1.
\end{matrix}\right.
\label{prior}
\end{equation}
This non-parametric formulation fits the mixture distribution without a constant number of components, allowing the mixture to dynamically adapt its cluster structure to the increased complexity of the lifelong learning process.

It may become an intractable combinatorial optimization problem to directly maximize the expected likelihood of the mixture model.
We need a scalable approximation that can represent the conditional distribution of the latent variable with maximum a posteriori (MAP) estimation.
Hence, we employ an expectation maximization (EM) procedure to update the mixture of task-specific parameters in an online manner, without access to samples from previous tasks.
Here, the E-step in EM computes the posterior expectation of task-to-cluster assignments, i.e., estimating the conditional mode of task-specific parameters, and the M-step involves updating parameters of all task models for future learning.

Let $p_{\bm{\theta}}(Y|X)$ denote the predictive likelihood function regarding the task model $\bm{\theta}$ on a batch of samples $(X,Y)$, i.e., $P(Y|X, \bm{\theta})$.
The predictive function treats each sample as an independent Gaussian $\mathcal{N}(y_i; \hat{y}_{\bm{\theta}}(x_i), \sigma^2)$ as
\begin{equation}
\begin{aligned}
p_{\bm{\theta}}(Y|X) & = \Pi_{i}\mathcal{N}(y_i; \hat{y}_i, \sigma^2) \\
& = \Pi_{i}\mathcal{N}\left(r_i+\gamma Q_{\bm{\varphi}}(s'_i,\mu_{\bm{\phi}}(s'_i)); Q_{\bm{\varphi}}(s_i,a_i), \sigma^2\right),
\label{pll}
\end{aligned}
\end{equation}
where $r_i\sim r(s_i, a_i)$ and $\sigma^2$ is a constant.
First, we estimate the expectation over pre-existing task clusters and the potential new one.
The posterior probability of task-to-cluster assignment $P(\bm{\theta}_t^{(l)}|X_t, Y_t)$ is calculated by the Bayes rule as
\begin{equation}
P(\bm{\theta}_t^{(l)}|X_t,Y_t) = \frac{P(Y_t|X_t, \bm{\theta}_t^{(l)})P(X_t|\bm{\theta}_t^{(l)})P(\bm{\theta}_t^{(l)})}{P(X_t, Y_t)}.
\label{ori_post}
\end{equation}
We assume that the input marginal likelihood $P(X_t|\bm{\theta}_t^{(l)})$ is approximately invariant across tasks and can be neglected.
Then, the posterior can be roughly approximated by
\begin{equation}
P(\bm{\theta}_t^{(l)}|X_t,Y_t) \varpropto p_{\bm{\theta}_t^{(l)}}(Y_t|X_t)P(\bm{\theta}_t^{(l)}).
\label{post}
\end{equation}
Combining the predictive likelihood in~(\ref{pll}) and the CRP prior distribution in~(\ref{prior}), we perform the E-step to compute the posterior probabilities of task-to-cluster assignments as
\begin{equation}
P(\bm{\theta}_t^{(l)}|X_t,Y_t) \varpropto \left\{\begin{matrix}
p_{\bm{\theta}_t^{(l)}}(Y_t|X_t)\sum_{t'=1}^{t-1}P(\bm{\theta}_{t'}^{(l)}), & l\le L, \\
p_{\bm{\theta}_t^{(l)}}(Y_t|X_t)\xi, & l=L+1.
\end{matrix}\right.
\label{posterior}
\end{equation}

With the estimated posterior task-to-cluster assignments, we turn to the M-step to maximize the expected log-likelihood of the mixture model as
\begin{equation}
\mathcal{L}(\bm{\theta}_t) = \mathbb{E}_{\bm{\theta}_t\sim P(\bm{\theta}_t|X_t, Y_t)}[\log p_{\bm{\theta}_t}(Y_t|X_t)].
\end{equation}
Supposing that each task model begins with some prior parameters $\bm{\theta}^-$, the value of $\bm{\theta}_t$ after taking all history gradient updates is calculated as
\begin{equation}
\begin{split}
& \bm{\theta}_{t+1}^{(l)} = \bm{\theta}_1^{(l)} + \\
& \alpha\sum_{t'=1}^{t}P(\bm{\theta}_{t'}^{(l)}|X_{t'},Y_{t'})\nabla_{\bm{\theta}_{t'}^{(l)}}\log p_{\bm{\theta}_{t'}^{(l)}}(Y_{t'}|X_{t'}),~\forall l,
\end{split}
\label{mstep}
\end{equation}
where $\alpha$ is the learning rate.
In the lifelong learning setting, all model parameters are updated at each time period.
Hence, we can perform the M-step in~(\ref{mstep}) by simply updating model parameters at the previous time period $\bm{\theta}_{t-1}$ on the newest samples as
\begin{equation}
\bm{\theta}_{t+1}^{(l)} = \bm{\theta}_{t}^{(l)} + \alpha P(\bm{\theta}_t^{(l)}|X_t,Y_t)\nabla_{\bm{\theta}_{t}^{(l)}}\log p_{\bm{\theta}_{t}^{(l)}}(Y_{t}|X_t),~\forall l.
\label{mstep2}
\end{equation}
This formation removes the requirement for memorizing samples of previous tasks, yielding a practical lifelong RL algorithm that tackles a continual stream of data.
Additionally, the E- and M-steps are iteratively alternated to converge to fully implement the EM algorithm.

\begin{algorithm}[tb]
	\caption{Scalable Lifelong RL with A Dirichlet Process Mixture}
	\label{llrl}
	\KwIn{Task sequence $\mathcal{D}=[\mathcal{D}_1,...,\mathcal{D}_{t-1},\mathcal{D}_t,...]$, \newline robust prior parameters $\bm{\theta}^-$}
	\KwOut{Optimal model parameters $\bm{\theta}_t^*$}
	Initialize $L=1, t=1, l^*=1$, and $\bm{\theta}_1^{(1)}\leftarrow\bm{\theta}^-$
	
	\For{each time period $t$}{
		Initialize $\bm{\theta}_t^{(L+1)}\leftarrow\bm{\theta}^-$ \\
		
		Receive a batch of transitions $\mathcal{T}_t\!=\!\sum_i(s_i,a_i,r_i,s'_i)$ \\ 
		
		Construct $(X_t,Y_t)$ from $\mathcal{T}_t$ \\
		
		Calculate $p_{\bm{\theta}_t^{(l)}}(Y_t|X_t)$ using~(\ref{pll}), $\forall l\le L+1$ \\
		
		Infer $P(\bm{\theta}_t^{(l)}|X_t,Y_t)$ using~(\ref{posterior}), $\forall l\le L+1$ \\
		
		\If{$P(\bm{\theta}_t^{(L+1)}|X_t,Y_t)\!>\!P(\bm{\theta}_t^{(l)}|X_t,Y_t), \forall l\le L$}{ 
			Add $\bm{\theta}_t^{(L+1)}$ to $\bm{\theta}_t$ thereafter \\
			
			$L\leftarrow L+1$
		}
		
		\While{not terminated}{
			E-step, re-calculate $P(\bm{\theta}_t^{(l)}|X_t,Y_t)$ using~(\ref{post}) with updated $\bm{\theta}_{t}^{(l)}$, $\forall l\le L$ \\
			
			M-step, adapt $\bm{\theta}_{t}^{(l)}$ using~(\ref{mstep2}) with updated $P(\bm{\theta}_t^{(l)}|X_t,Y_t)$, $\forall l\le L$ \\
		}
		
		$\bm{\theta}_{t+1}^{(l)}\leftarrow \bm{\theta}_{t}^{(l)}$, $\forall l\le L$ \\
		
		$l^* = \argmax_{l\le L}p_{\bm{\theta}_{t+1}^{(l)}}(Y_t|X_t)$ \\
	}
\end{algorithm}

We summarize the proposed lifelong RL algorithm and outline it in Algorithm~\ref{llrl}.
At the nominal time period, we initialize the mixture model that contains one entry $\bm{\theta}_1^{(1)}\leftarrow\bm{\theta}^-$ in Line~$1$.
From there, at each time step $t$, we first initialize an empty task model from the prior parameters $\bm{\theta}^-$ in Line~$3$.
Then, we collect a batch of transitions in Line~$4$ and construct the input-output samples in Line~$5$.
Next, we calculate the predictive likelihood of all task models in Line~$6$, and infer the posterior task-to-cluster assignments in Line~$7$.
The CRP prior allows some probability for instantiating a new cluster to the mixture distribution, while the posterior task-to-cluster assignments determine the expansion of the new task cluster into the mixture model.
In Lines~$8$-$11$, the new potential cluster is added to the mixture if its posterior probability is larger than those of the $L$ pre-existing clusters.
Then, we keep alternating the E- and M-steps until the learning is terminated in Lines~$12$-$15$, \footnote{Empirically, the learning is terminated when the change of the neural network weights $\bm{\theta}$ is smaller than a preset tiny threshold.} 
and obtain the updated model parameters in Line~$16$.
Using updated model parameters $\bm{\theta}_{t+1}^{(l)}$, the next batch of transitions is predicted according to the most likely task $l^*$ in Line~$17$.

\subsection{Robust Prior via Domain Randomization}\label{dr}
We formulate a mixture of task models for performing lifelong learning adaptation in the face of an infinite stream of incoming data.
New task models are instantiated as needed under the Bayesian inference framework, where parameters of each new task model are initialized from $\bm{\theta}^-$.
However, modern parametric models, e.g., DNNs,  are usually hard to train in such a lifelong learning setting.
They typically require numerous iterations with plenty of training samples to learn a sensible solution, which can be infeasible when faced with lifelong streaming information.
Therefore, we employ the \textit{domain randomization} approach~\citep{tobin2017domain,muratore2019assessing} to train the prior parameters $\bm{\theta}^-$ for each task model.
Domain randomization is originally proposed to learn control policies \textit{robust} to the transfer from simulation to reality, i.e., ``sim-to-real", by randomizing various aspects of the simulated environment at training time.
We adopt this technique to learn robust model initialization that can generalize well to non-stationary task distributions during lifelong learning.

In contrast to learning a policy for one particular task, we train a model $\bm{\theta}^-$ that is capable of tackling a diversity of tasks.
In the lifelong learning setting, we collect samples from a finite number of tasks $\mathcal{D}^-=[\mathcal{D}_1,...,\mathcal{D}_m]$, and use all these samples to train a robust model prior.
The objective is then modified to minimize the expected loss, i.e., the Bellman residual, across a distribution of tasks as
\begin{equation}
\begin{aligned}
&\mathcal{L}(\bm{\theta}^-)=\mathbb{E}_{\mathcal{D}_i\sim\mathcal{D}^-}\left[\mathbb{E}_{x,y\sim\mathcal{D}_i}\left[(\hat{y}- y)^2\right]\right]=\\ &\mathbb{E}_{\mathcal{D}_i\sim\mathcal{D}^-}\!\!\left[\mathbb{E}_{s,a,r,s'\sim\mathcal{D}_i}\!\!\left[\!(r\!+\!\gamma Q_{\bm{\varphi}^-}(s',\mu_{\bm{\phi}^-}(s'))\!-\!Q_{\bm{\varphi}^-}(s,\!a))^2 \right]\right]\!\!.
\end{aligned}
\end{equation}
By training the model to adapt to variability in the non-stationary task distribution, the resulting model is supposed to better generalize to unseen tasks.
After a new task model is instantiated from the prior parameters $\bm{\theta}^-$, it might then better adapt to any task using only a few gradient steps.

\section{Experiments}\label{experiments}
Experiments are conducted on a suite of continuous control tasks to show the applicability and scalability of our method in lifelong learning settings.
Using agents in these tasks, we create a variety of representative RL problems with non-stationary task distributions, where scalable lifelong learning is crucial.
The following two subsections show empirical results and corresponding insights on the experimentation.
We compare our method to several baseline methods:
\begin{enumerate}
	\item \textbf{Fine-tune}. As a representative dynamic evaluation baseline in the literature~\citep{krause2018dynamic,nagabandi2019learning}, it continually trains a single base model as the steaming data enters.
	\item \textbf{Reservoir}. Simple experience replay with reservoir sampling can be a powerful tool in lifelong learning~\citep{chaudhry2019continual,rolnick2019experience}. 
	It uniformly selects a batch of samples from an infinite data stream, which is well suited to manage the replay memory without explicit task boundaries.
	\item \textbf{Consolidation}. The policy consolidation~\citep{kaplanis2019policy} method uses a cascade of hidden networks to simultaneously remember policies at a range of timescales and regularize the current policy by its own history~\citep{kaplanis2018continual}. 
	Since it does not require knowledge of task boundaries, we evaluate this regularization-based method for comparison.
	\item \textbf{Progressive}. We evaluate progressive neural network~\citep{rusu2016progressive} as a classical expansion-based baseline for lifelong learning, which freezes the previous network and allocates new sub-networks to accommodate new information~\citep{yoon2018lifelong}.
	Note that it requires explicit task boundaries and labels.
\end{enumerate}
 
We use the DDPG algorithm to handle continuous control tasks, where the actor maps a given state to an estimated optimal action and the critic approximates the action-value function.
Both the actor and critic are represented by a neural network containing two 512-node hidden layers with ReLU activation, and their parameters are optimized by gradient descent.
To promote good stability~\citep{lillicrap2016continuous}, we utilize the soft updating strategy to update weights of target networks.
For Reservoir, the model is fed with two mini-batches of the same size at each learning iteration: one from the current data stream and the other from the long-term replay memory.
For Consolidation, the model consists of $4$ sets of hidden networks in addition to the visible one that is amenable for the current policy.
The hyper-parameters are set as: learning rate $\alpha=0.001$, discounting factor $\gamma=0.99$, batch size for network updating $n=64$.

We define two performance metrics for each evaluation unit, i.e., a given tested approach running on a given task.
One is the return of one learning episode that is associated with the learned policy, defined as $\sum_{i=1}^Hr(s_i, \mu_{\bm{\phi}}(s_i))$, where $H$ is the time horizon of the learning episode.
The other is the average return over all learning episodes, defined as $\frac{1}{J}\sum_{j=1}^J\sum_{i=1}^Hr(s_i^j, \mu_{\bm{\phi}}(s_i^j))$, where $J$ is the number of learning episodes.
The former will be plotted in figures and the latter will be presented in tables.
We continually change the learning task at random for $T=50$ times to create a lifelong learning process with a non-stationary task distribution $\mathcal{D}=[\mathcal{D}_1,...,\mathcal{D}_T]$.
We report the performance of all tested approaches for each task instance $\mathcal{D}_t (1\le t\le T)$, and record the statistical outcome over all encountered tasks to demonstrate the capability of lifelong learning. 
Our code is available online.
\footnote{\url{https://github.com/HeyuanMingong/sllrl}}

\subsection{Simple 2D Navigation}
\begin{figure}[tb]
	\centering
	\includegraphics[width=0.95\columnwidth]{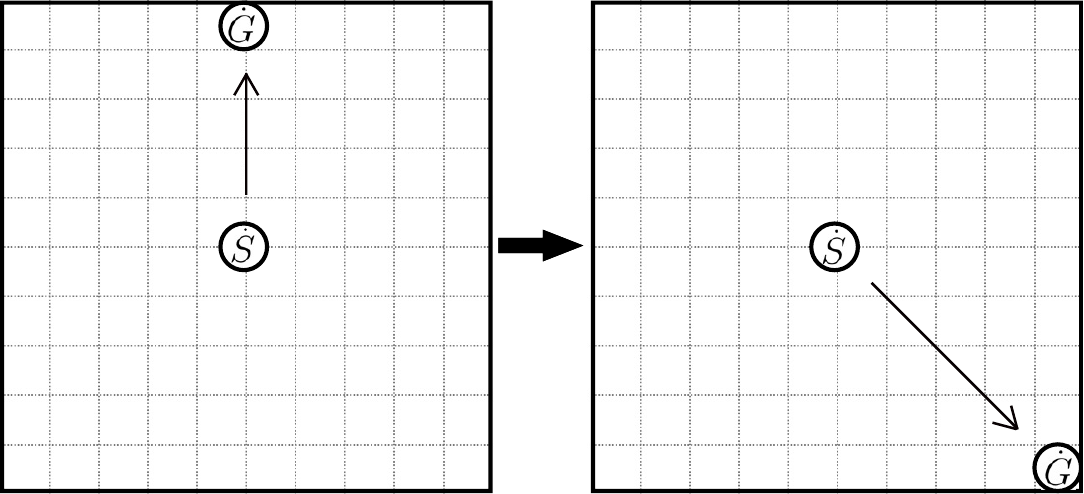}
	\caption{The simple 2D navigation task in a lifelong learning setting where the goal may change over time. $\dot{S}$ is the starting point and $\dot{G}$ is the goal.}
	\label{fig:navi}
\end{figure}

As an explanatory experiment, we implement a simple 2D navigation task in a continuous state-action space to serve as a proof of principle and test if our method achieves both plasticity and stability in lifelong learning scenarios.
As shown in Fig.~\ref{fig:navi}, the task is to move a point agent to a goal position within a unit square.
The state is the agent's current position in the 2D coordination system.
The action is a 2D velocity vector and is clipped to the range of $[-0.1, 0.1]$.
The reward function is set as the negative Euclidean distance to the goal position minus a minor control cost proportional to the action magnitude.
Each learning episode begins with a fixed initial state, and terminates when the point agent reaches the region within $0.01$ of the goal position or the time step comes to the horizon of $H=100$.
During lifelong learning, the goal position may change over time within the unit square at random, resulting in a non-stationary task distribution.

\begin{figure}[tb]
	\centering
	\includegraphics[width=0.95\columnwidth]{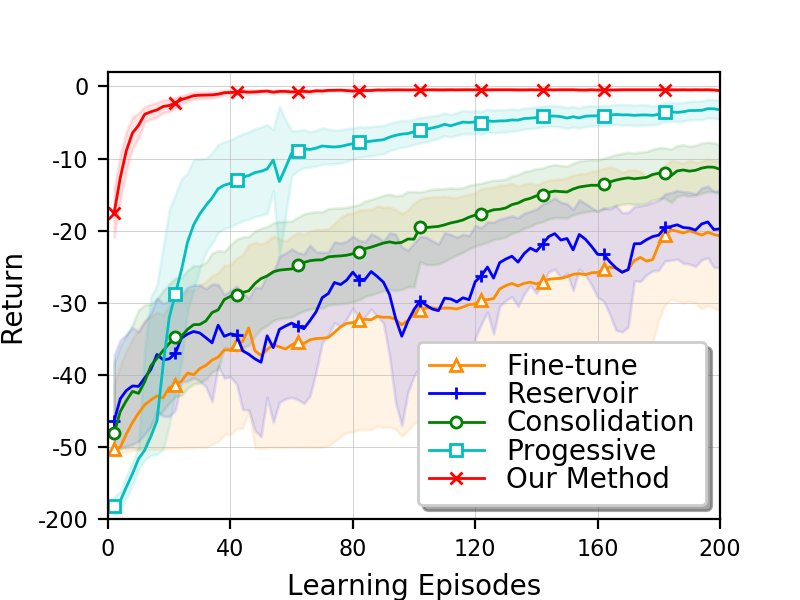}
	\caption{The return per learning episode of baselines and our method in the 2D navigation task. 
	Here and in similar figures below, the bold line depicts the mean of received return per episode over $T=50$ sequential tasks, and the shaded plots $95\%$ bootstrapped confidence intervals of the mean.}
	\label{fig:navi-rews}
\end{figure}

\begin{figure*}[tb]
	\centering
	\subfigure[$t=3$]{\includegraphics[width=0.3\textwidth]{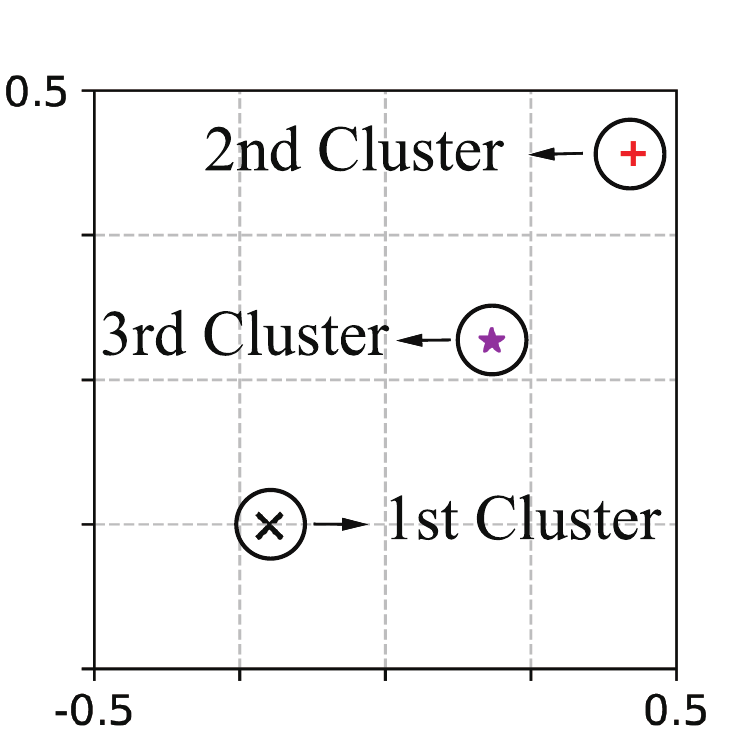}}
	\subfigure[$t=16$]{\includegraphics[width=0.3\textwidth]{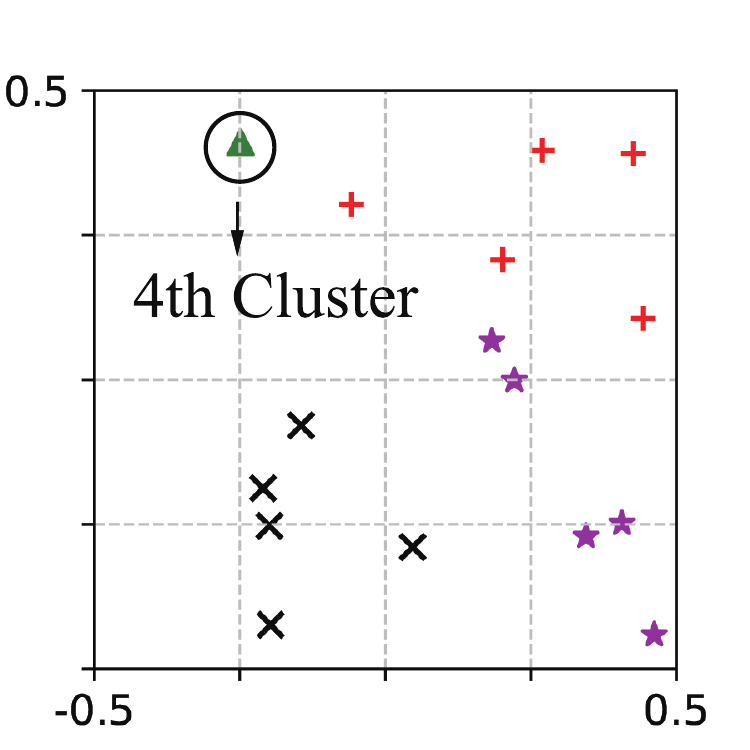}}
	\subfigure[$t=T (50)$]{\includegraphics[width=0.3\textwidth]{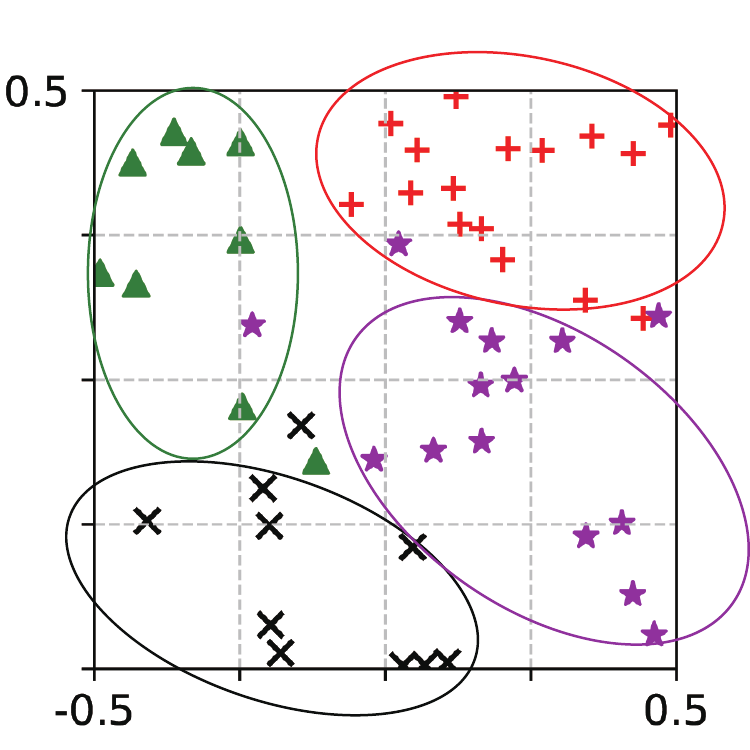}}
	\caption{A visualization of the Bayesian mixture during lifelong learning: (a) The initial $3$ clusters are instantiated at the initial $3$ time steps; (b) The 4th cluster is instantiated at time step $t=16$; (c) All the $T=50$ tasks are clustered into $4$ mixture components effectively.}
	\label{fig:c4}
\end{figure*}

We first show main results of our method and baseline methods implemented on the 2D navigation task.
For our method, $L=4$ task clusters are instantiated totally.
Fig.~\ref{fig:navi-rews} presents the received return per learning episode, and Table~\ref{table:navi_rews} reports the numerical average return over 200 learning episodes.
Fine-tune obtains the worst lifelong learning performance under a non-stationary task distribution, since it adopts the simplest learning adaptation mechanism.
Reservoir achieves slightly higher average return than Fine-tune, while its received return per learning episode tends to oscillate more during lifelong learning.
We conjecture that replaying samples from other tasks can impose some interference on the online updates when learning a new task.
Consolidation obtains better performance than Reservoir, indicating that regularizing the current policy by its own history to force it less overfitted to the task at hand can alleviate catastrophic interference to some extent.
Progressive performs the best among the baseline methods, which is supposed to benefit from maintaining stability by blocking changes to the previous network and promoting plasticity by allocating new sub-networks to accommodate new knowledge.

\begin{table}[tb]
	\caption{Numerical average return over $200$ episodes of baselines and our method on the 2D navigation task. 
	Here and in similar tables below, we present the mean over $T=50$ sequential tasks and corresponding standard errors.
	We mark the best performance in boldface.}
	\centering
	\setlength{\tabcolsep}{5mm}\renewcommand\arraystretch{1.2}
	\begin{tabular}{c|c}
		\cmidrule[\heavyrulewidth]{1-2}
		Methods     	&    	Return \\
		\hline 
		Fine-tune     	&  		$-31.73\pm 0.51$ \\
		Reservoir 		&  		$-28.78\pm 0.48$ \\
		Consolidation 	&  		$-22.01\pm 0.64$ \\
		Progressive 	& 		$-15.47\pm 2.07$ \\
		Our Method & 		$\bm{-1.23\pm 0.17}$ \\
		\cmidrule[\heavyrulewidth]{1-2}
	\end{tabular}
	\label{table:navi_rews}
\end{table}

In contrast, it can be observed from Fig.~\ref{fig:navi-rews} that our method achieves much faster learning adaptation to the non-stationary task distribution compared to all baseline methods. 
Our method takes only $20$ learning episodes to obtain near-optimal asymptotic performance for a given task during lifelong learning, while it takes far more than $100$ episodes for all baseline methods to achieve comparable performance.
Table~\ref{table:navi_rews} illustrates that our method obtains remarkably greater average return over all learning episodes than all baselines. 
Based on explicitly estimating task relatedness, our method is capable of enhancing stability by modulating transferability across tasks and promoting plasticity by recognizing outlier tasks that require a more significant degree of adaptation.
Moreover, statistical results show that our method achieves narrower confidence intervals and smaller standard errors than all baselines, indicating that our method enables more stable lifelong learning adaptation to a changing distribution of tasks.

Furthermore, to test whether our method correctly estimates task relatedness and clusters encountered tasks in a latent space, we gain an intuition of the Bayesian mixture via visualization to observe and comprehend the lifelong learning process.
Each task is characterized by the reward function associated with its goal position.
Tasks with adjacent goals reveal higher similarity and are likely to be assigned to the same mixture component.
Therefore, we employ the goal position in the 2D coordinate system as a visualization to measure relatedness between tasks.
As illustrated in Fig.~\ref{fig:c4}, each data point within the unit square represents a goal position associated with a particular task $\mathcal{D}_t (1\le t\le T)$.
Tasks belonging to different clusters in the mixture are depicted by data points with different colors and shapes.
We can observe that the four task clusters are expanded into the mixture model at time steps $t=1,2,3,16$ incrementally.
During the entire lifelong learning process, the tasks under a non-stationary distribution over $T=50$ time steps are consecutively clustered as four mixture components in a latent space, as visualized in Fig.~\ref{fig:c4}-(c).
It successfully verifies that our method is capable of clustering tasks from a non-stationary distribution in a latent space where similar tasks are closely spaced and tend to be assigned to the same cluster.
This is crucial for a scalable lifelong RL algorithm since correctly estimating task relatedness is the prerequisite for modulating transferability across tasks.
At each time step, the task at hand is allocated to a pre-existing cluster or expanded as a new component in the mixture according to the non-parametric Bayesian framework.
This non-parametric formulation fits the mixture distribution without a priori fixed number of components and without any external information to signal task boundaries in advance, which is critical for scalable lifelong learning in real world.

\subsection{MuJoCo Locomotion}
The results in the simple 2D navigation domain demonstrate that our method facilitates scalable lifelong RL with good balance between stability and plasticity.
Next, we test whether similar benefits can be obtained for lifelong learning when our method is applied to more sophisticated deep RL problems at the scale of DNNs.
In the next set of experiments, we investigate two kinds of high-dimensional continuous control problems based on the MuJoCo physics engine~\citep{todorov2012mujoco}.

\begin{figure}[tb]
	\centering 
	\subfigure[Reacher]{\includegraphics[width=0.45\columnwidth]{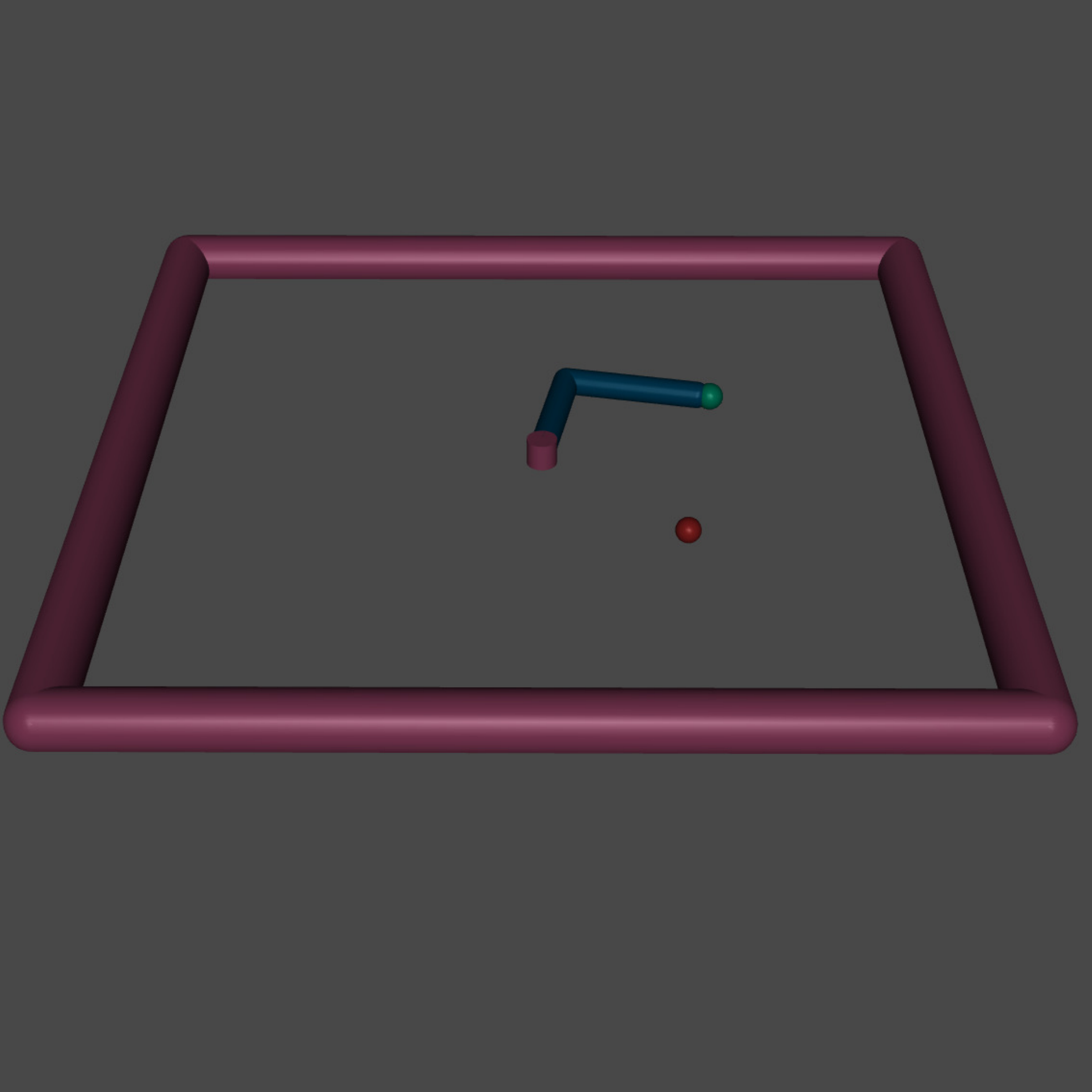}}
	\subfigure[Hopper]{\includegraphics[width=0.45\columnwidth]{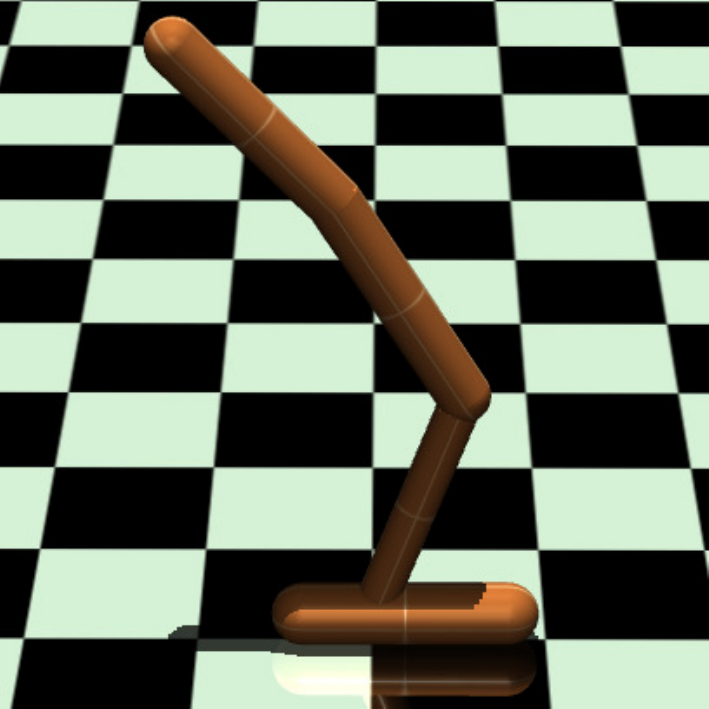}}
	\caption{Two kinds of MuJoCo locomotion tasks: (a) Reacher, (b) Hopper.}
	\label{fig:envs}
\end{figure}

As shown in Fig.~\ref{fig:envs}-(a), one is the Reacher domain that aims to move a two-joint torque-controlled robot arm to a particular target point.
The reward function is set as the negative Euclidean distance between the fingertip and the target location, minus a minor control penalty proportional to the scale of action.
Each learning episode begins with a fixed initial state, and terminates when the fingertip reaches the region within $0.001$ of the target location or the time step comes to the horizon of $H=100$.
The lifelong task distribution is created by changing the target point within the reachable circle at random. 
As shown in Fig.~\ref{fig:envs}-(b), the other is the Hopper domain that requires a one-legged hopper robot to run forward at a given velocity along the $x$-axis.
The reward function is set as the negative absolute difference between the current velocity of the robot and a goal one, plus an alive bonus.
Each learning episode terminates when the robot falls down or the time step comes to the horizon of $H=100$.
We consecutively change the goal velocity randomly within the range of $[0.0, 1.0]$, resulting in a non-stationary task distribution.

\begin{table}[tb]
	\caption{Numerical average return over $200$ learning episodes of baselines and our method implemented on the two kinds of MuJoCo locomotion domains.}
	\centering
	\setlength{\tabcolsep}{5.0mm}\renewcommand\arraystretch{1.2}
	\begin{tabular}{c|c|c}
		\cmidrule[\heavyrulewidth]{1-3}
		Methods 		& Reacher & Hopper \\
		\hline 
		Fine-tune    		      & $-8.58\pm 0.20$          & $-4.56\pm 0.08$ \\
		Reservoir 			       & $-4.22\pm 0.07$          & $-4.71\pm 0.02$ \\
		Consolidation 		    & $ -9.18\pm 0.19$         & $-4.36\pm 0.08$ \\
		Progressive 		    & $ -4.30\pm 0.18$         & $-2.62\pm 0.15$ \\
		Our Method 	 & $\bm{-2.63\pm 0.11}$ & $\bm{0.05\pm 0.11}$ \\
		\cmidrule[\heavyrulewidth]{1-3}
	\end{tabular}
	\label{table:mujoco_rews}
\end{table}

\begin{figure}[tb]
	\centering
	\subfigure[Reacher]{\includegraphics[width=0.95\columnwidth]{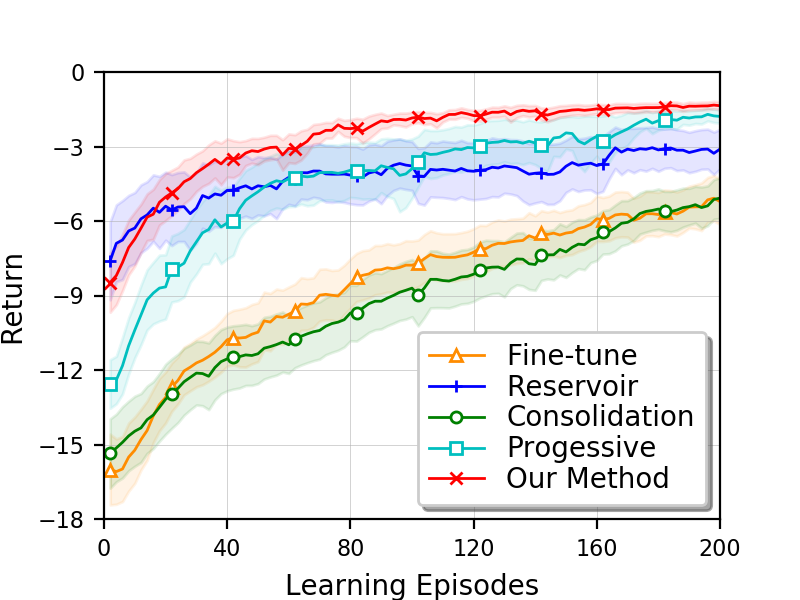}}
	\subfigure[Hopper]{\includegraphics[width=0.95\columnwidth]{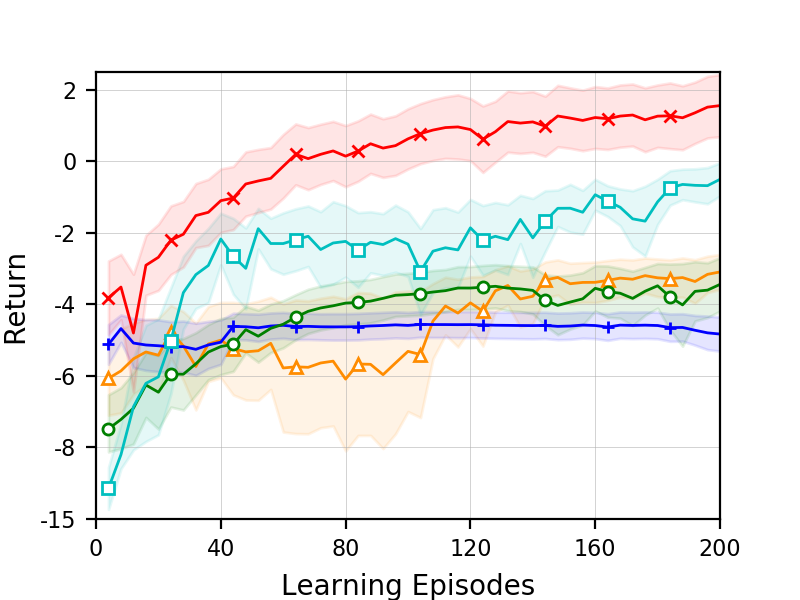}}
	\caption{The received return per learning episode of baselines and our method in the two kinds of MuJoCo locomotion domains: (a) Reacher, (b) Hopper.}
	\label{fig:mujoco-rews}
\end{figure}

We show primary results of baselines and our method implemented on the two kinds of locomotion domains.
Fig.~\ref{fig:mujoco-rews} illustrates the received return per episode, and Table~\ref{table:mujoco_rews} presents corresponding numerical average return over $200$ learning episodes.
Fine-tune and Consolidation obtain similar performance in both domains.
Reservoir can receive high returns at the early learning stage, while it tends to achieve sub-optimal performance later on since replaying old samples of previous tasks may interfere with learning the new task.
Progressive also performs better than other baselines, demonstrating the effectiveness and superiority of expansion-based approaches for lifelong learning.
Our method usually achieves significantly more rapid and stable lifelong learning adaptation compared to baseline approaches.
By comparison, it consumes significantly more computation cost for baseline approaches to achieve comparable performance to our method.
For example, in the Hopper domain, our method only needs approximately $40$ learning episodes to achieve a near-optimal return, while all baselines can cost far more than $200$ episodes.

The results reveal that our method effectively builds on previously learned knowledge to improve learning adaptation to new tasks throughout the lifetime.
Governed by the Bayesian non-parametric framework, the task identity at each time period is automatically detected by MAP estimation.
Subsequently, our method retrieves the most similar experience from the mixture of robust task models (including the potential new cluster), which is supposed to benefit the new task at hand most.
By modulating transferability across tasks, our method only requires to ``fine-tune" the selected prior experience a little bit using a small quantity of computational efforts, being significantly more efficient for lifelong learning adaptation to non-stationary task distributions.

\begin{figure*}[tb]
\centering
\subfigure[Navigation]{\includegraphics[width=0.32\textwidth]{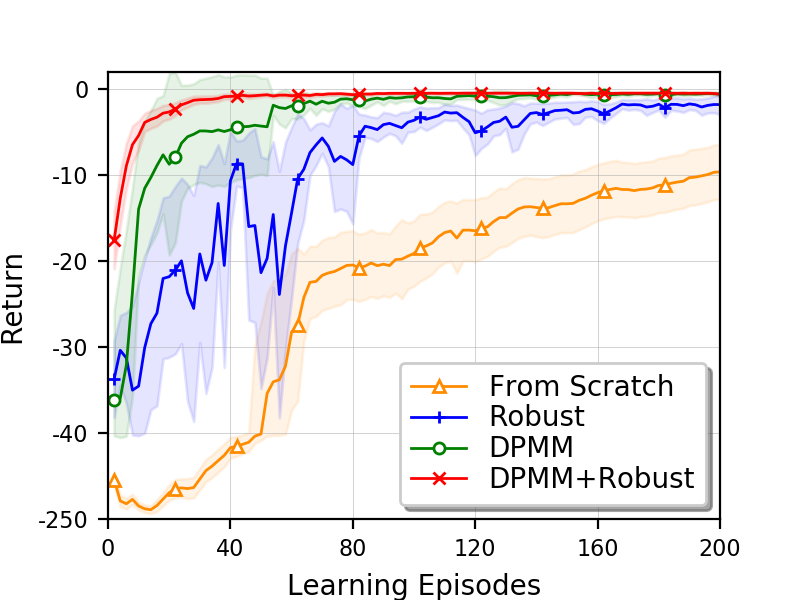}}
\subfigure[Reacher]{\includegraphics[width=0.32\textwidth]{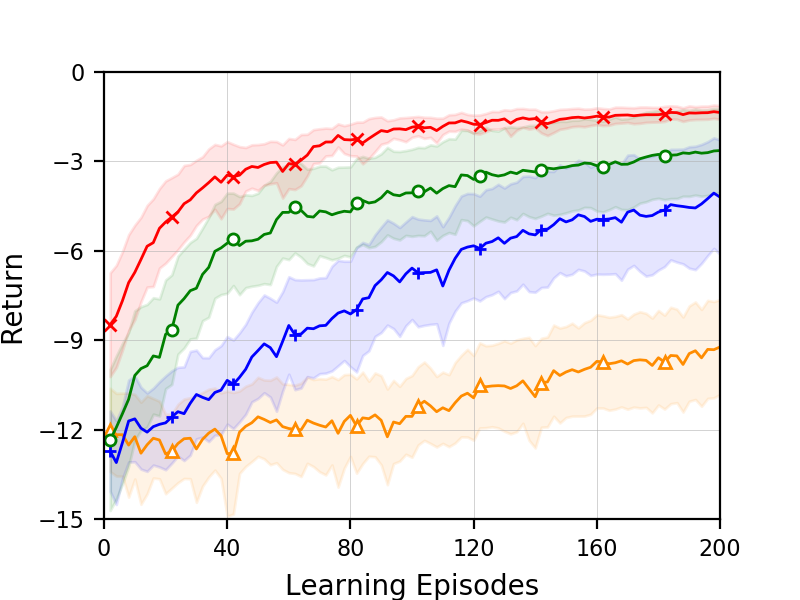}}
\subfigure[Hopper]{\includegraphics[width=0.32\textwidth]{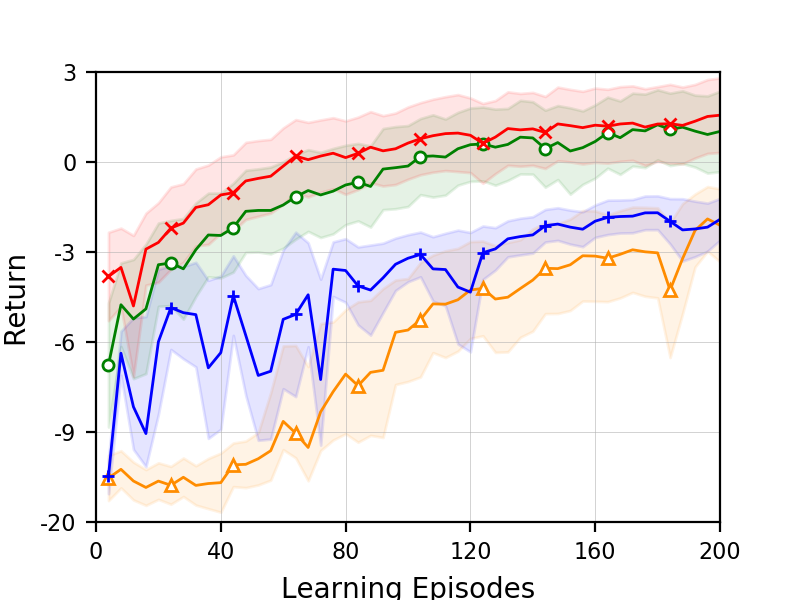}}
\caption{The received return per learning episode of the ablation study on all domains: (a) Navigation, (b) Reacher, (c) Hopper.}
\label{fig:ablation}
\end{figure*}

\subsection{Ablation Study}
To identify the respective contribution of the two components to overall performance, i.e., DPMM and domain randomization, we conduct an ablation study to separate the two components apart for observation on both the Navigation and MuJoCo domains.
During lifelong learning, we implement four variants of our method as follows.
\begin{enumerate}
	\item \textbf{From Scratch.} Without component used, it learns each task from scratch, providing a lower bound to show the benefits of lifelong transfer in general.
	\item \textbf{Robust.} We employ domain randomization to train a robust prior, and learn each task using DDPG with parameters initialized from that prior.
	\item \textbf{DPMM.} The domain randomization component is ablated from our method.
	\item \textbf{DPMM+Robust.} Both components are used.
\end{enumerate}
Fig.~\ref{fig:ablation} shows the received return per episode, and Table~\ref{table:ablation} presents numerical average return over 200 learning episodes.

First, we identify how DPMM affects the lifelong learning performance by comparing From Scratch with DPMM, and by comparing Robust with DPMM+Robust.
It is observed that DPMM and DPMM+Robust can largely improve the performance of From Scratch and Robust, respectively, which verifies the significant effectiveness of our DPMM component.
With formulating the non-stationary task distribution with an increasing number of clusters, our DPMM component provides a flexible structure for modulating transferability across tasks and accommodating new knowledge as needed.

Next, we test the capability of the adopted domain randomization technique.
Comparing From Scratch with Robust, we can observe that domain randomization is capable of improving learning performance to a large extent.
It validates the promising efficiency of domain randomization for learning robust model initialization that can generalize well to non-stationary task distributions during lifelong learning.
Comparing DPMM with DPMM+Robust, it is observed that DPMM can achieve a moderate performance improvement with the help of domain randomization, as the promotion only occurs when a new task cluster is expanded into the mixture.

Finally, we compare all the four variants.
DPMM is the crucial component of our method, in that removing this component can cause a large drop in learning performance.
DPMM better facilitates learning performance than domain randomization, and combining the two components jointly results in the best lifelong learning adaptation to the non-stationary task distribution.

\begin{table}[tb]
\caption{Numerical average return over $200$ learning episodes of the ablation study on all domains.}
\centering
\setlength{\tabcolsep}{1.5mm}\renewcommand\arraystretch{1.2}
\begin{tabular}{c|c|c|c}
	\cmidrule[\heavyrulewidth]{1-4}
	Methods 		&	Navigation & Reacher & Hopper \\
	\hline 
	From Scratch 	& $-49.79\pm 4.61$ & $-11.12\pm 0.08$ & $-7.50\pm 0.33$ \\
	Robust 			& $-9.08\pm 0.66$  & $-7.45\pm 0.18$ & $-4.14\pm 0.18$\\
	DPMM 			& $-3.37\pm 0.46$  & $-4.78\pm 0.16$ & $-0.75\pm 0.14$ \\
	DPMM+Robust 	& $\bm{-1.23\pm 0.17}$ & $\bm{-2.63\pm 0.11}$ & $\bm{0.05\pm 0.11}$ \\
	\cmidrule[\heavyrulewidth]{1-4}
\end{tabular}
\label{table:ablation}
\end{table}

\section{Related Work}\label{related}
Lifelong learning considers learning multiple tasks in sequence,
which needs to retain previously learned knowledge and leverage that knowledge to facilitate learning new skills~\citep{lopez2017gradient}.
Various configurations in the literature are related to lifelong RL.
Multi-task RL~\citep{hessel2019multi} aims to optimize the overall performance of all tasks, which needs a reservoir of persistent training samples for all tasks.
Transfer RL~\citep{pan2018multisource,yang2020efficient} assumes the simultaneous availability of multiple source tasks and concentrates on facilitating the performance of a particular target task.
Meta-RL~\citep{finn2017model,nagabandi2019learning}, also called as few-shot RL, learns a base model (i.e., the meta) that can quickly adapt to new tasks, while not considering the alleviation of catastrophic forgetting or interference.

A variety of approaches have been investigated to tackle catastrophic forgetting or interference in the machine learning community.
These can be classified into three major categories according to how the knowledge of previous tasks is memorized and leveraged: replay-based, regularization-based, and expansion-based.

Replay-based approaches use the idea of episodic memory, where examples from prior tasks are stored to recall experiences encountered in the past.
While storing past examples for rehearsal can date back to 1990s~\citep{robins1995catastrophic}, it yields decent results against catastrophic forgetting in practical problems.
\citet{rolnick2019experience} leveraged off-policy learning from replay experiences to enhance stability, and used behavior cloning to keep the policy distribution close to historical data.
\citet{isele2018selective} proposed a rank-based method for the online collection and preservation of training experiences in a long-term memory to reduce the effects of forgetting.
Instead of storing training examples, \citet{lopez2017gradient} stored gradients of previous tasks, such that at any time the gradients of all tasks except the current one can be used to form a trust region that prevents forgetting.
An inherent drawback is the constraint on the memory capacity as the number of encountered tasks grows, which could limit its application to large-scale problems. 
To avoid storing past examples, \citet{shin2017continual} sampled synthetic data from a generative model, shifting the problem to the training of this generative model.
However, the generative model used to mimic older parts of the data distribution can also suffer from catastrophic forgetting~\citep{kaplanis2019policy}.
Furthermore, the policy trained using experiences from an enormous range of domains may learn a conservative strategy or fail to learn the task~\citep{yu2019policy}.

Regularization-based approaches are typically inspired by theoretical neuroscience suggesting that synapses with different levels of plasticity can protect consolidated knowledge from forgetting~\citep{benna2016computational}.
From a computational perspective, additional regularization terms are imposed on the learning objective, aiming to identify the important weights of previous tasks and penalize large updates on those weights when learning a new task.
Elastic weight consolidation (EWC)~\citep{kirkpatrick2017overcoming} slowed down the learning for weights relevant to knowledge of previous tasks by adding a quadratic penalty on the difference between parameters of the old and new tasks weighted by the Fisher information matrix.
Similar to EWC, \citet{zenke2017continual} maintained an online estimate of the synapse's importance regarding past tasks and penalized changes to the most relevant synapses, such that new tasks are trained with minimal forgetting.
\citet{schwarz2018progress} used a modified version of EWC to mitigate forgetting when distilling the newly learned behavior into the knowledge base.
\citet{kaplanis2018continual,kaplanis2019policy} proposed a cascade of hidden networks that simultaneously remember policies at a range of timescales and regularized the current policy by its own history, thereby improving its ability to learn without forgetting.
In general, with limited neural resources, comprising additional regularization terms may lead to a trade-off on the accomplishment of old and new tasks~\citep{parisi2019continual}.

On the other hand, expansion-based approaches incrementally expand new architectural resources, e.g., a policy/option library or the network capacity, in response to new information.
Conceptually, such a direction has two superiorities compared with the above two: (i) catastrophic forgetting is mitigated by protecting past memories from being perturbed by the new information; (ii) the model capacity is determined adaptively throughout the lifetime.
The family of policy reuse algorithms~\citep{fernandez2006probabilistic,fernandez2010probabilistic,glatt2017policy,glatt2020decaf} improved its exploration in a new task by probabilistic exploitation of similar policies from a built policy library.
Analogously, option reuse approaches~\citep{brunskill2014pac,bonini2018framework} summarized prior experience through temporally extended actions (i.e., sub-policies or options) and leveraged only reusable parts of the policy for future learning.
Another way is to expand the neural network capacity in the context of deep learning. 
The simplest example is to freeze early layers and fine-tune later layers when learning the new task~\citep{oquab2014learning}.
\citet{rusu2016progressive,rusu2017sim} blocked any changes to the network trained on previous tasks and allocated a new sub-network with fixed capacity to process the new information.
Similarly, dynamically expanding network~\citep{yoon2018lifelong} increased the amount of trainable parameters to accommodate new tasks incrementally and used group sparse regularization to decide how many neurons to add at each layer.
\citet{parisi2017lifelong} used self-organizing networks to update connectivity patterns and allocate neural resources dynamically for lifelong learning of human action sequences.

Nevertheless, existing expansion-based approaches usually suffer from the lack of scalability due to two critical limitations: (i) most of them are studied in a rather restricted setting that requires explicit task boundaries and hand-designed heuristics for incorporating new resources; (ii) the network size may scale quadratically in the number of encountered tasks.
In contrast, we use a Dirichlet process mixture to handle the non-stationary task distribution and automatically infer task identities under the Bayesian non-parametric framework, thereby achieving scalable lifelong RL.
The proposed method is an extension of our previous work in~\citep{wang2020lifelong}, which requires an auxiliary set of networks to approximate the reward or state transition function.
In this paper, we capture task relatedness using Bayesian inference on the Bellman residual, thus introducing only a single set of networks to concurrently train the policy and parameterize the task.

\section{Conclusions and Future Work}\label{conclusions}
In the paper, we propose a scalable lifelong RL method that dynamically expands the network capacity to quickly accommodate new knowledge while stably preserving past memories.
The non-stationary task distribution is modeled by a Dirichlet process mixture that clusters the task-specific parameters in a latent space.
Governed by the Bayesian non-parametric framework, the mixture is maintained via an EM procedure, in conjunction with a CRP prior, to dynamically adapt the model complexity without explicit task boundaries or hand-designed heuristics.
Based on capturing task relatedness by estimating the likelihood of task-to-cluster assignments, our method successfully enhances stability by modulating transferability across tasks, and promotes plasticity by recognizing outlier tasks that require a more significant degree of adaptation.
Furthermore, the domain randomization technique is employed to train robust task models for initializing the mixture components, thereby providing better generalization ability when adapting to unseen tasks.
Experiments conducted on a suite of continuous control domains verify that our method facilitates scalable lifelong learning performance to non-stationary task distributions.

A few interesting research directions are worth investigating for future work.
One is to evaluate our method on different domains, such as Atari games~\citep{mnih2015human} and StarCraft II learning environment~\citep{vinyals2019grandmaster}.
Another is to improve the accuracy of task inference, which is the main bottleneck of our method and could be addressed from several aspects.
For example, task relatedness is captured using Bayesian inference on the Bellman residual, where the ``pseudo" ground truth relies on the Q-network and will gradually change as the learning proceeds, analogous to the classical DQN algorithm~\citep{mnih2015human}.
To better capture task relatedness and modulate task transferability, we could use powerful variational inference approaches~\citep{hoffman2013stochastic} to more accurately approximate posterior distributions of task-to-cluster assignments. 
For another example, we neglect the input marginal likelihood in (\ref{ori_post}) for simplifying the posterior derivation.
We could employ efficient density estimators, e.g., VAE~\citep{kingma2014auto}, to describe the marginal likelihood for more accurate posterior inference.

\footnotesize
\bibliography{sllrl}
\bibliographystyle{IEEEtranN}

\begin{IEEEbiography}[{\includegraphics[width=1in,height=1.25in,clip,keepaspectratio]{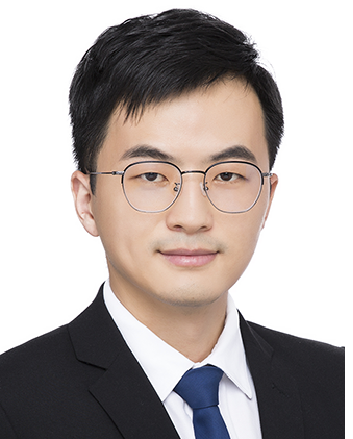}}]{Zhi Wang}
	(Member, IEEE) received the Ph.D. degree in machine learning from the Department of Systems Engineering and Engineering Management, City University of Hong Kong, Hong Kong, China, in 2019, and the B.E. degree in automation from Nanjing University, Nanjing, China, in 2015.
	He is currently an Assistant Professor with the School of Management and Engineering, Nanjing University, Nanjing, China.
	He had the visiting position at the University of New South Wales, Canberra, Australia, and holds the visiting position at the State Key Laboratory of Management and Control for Complex Systems, Institute of Automation, Chinese Academy of Sciences, China.
	
	His current research interests include reinforcement learning, machine learning, and robotics.
\end{IEEEbiography}

\begin{IEEEbiography}[{\includegraphics[width=1.0in,height=1.25in,clip,keepaspectratio]{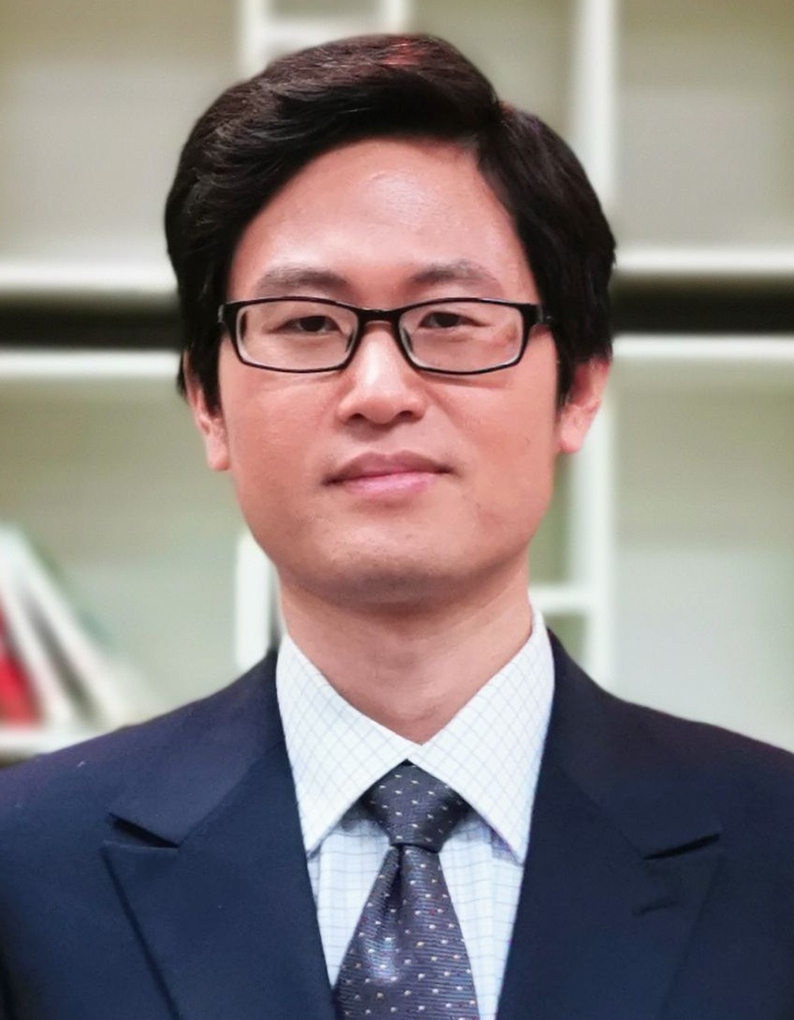}}]{Chunlin Chen}
	(Senior Member, IEEE) received the B.E. degree in automatic control and the Ph.D. degree in control science and engineering from the University of Science and Technology of China, Hefei, China, in 2001 and 2006, respectively.
	
	He was with the Department of Chemistry, Princeton University, Princeton, NJ, USA, from September 2012 to September 2013. He had visiting positions at the University of New South Wales, Canberra, ACT, Australia, and the City University of Hong Kong, Hong Kong. 
	He is currently a Professor and Vice Dean with the School of Management and Engineering, Nanjing University, Nanjing, China.
	His current research interests include machine learning, intelligent control, and quantum control.
	
	Dr. Chen serves as the Chair for the Technical Committee on Quantum Cybernetics, IEEE Systems, Man and Cybernetics Society.
\end{IEEEbiography}

\begin{IEEEbiography}[{\includegraphics[width=1.0in,height=1.25in,clip,keepaspectratio]{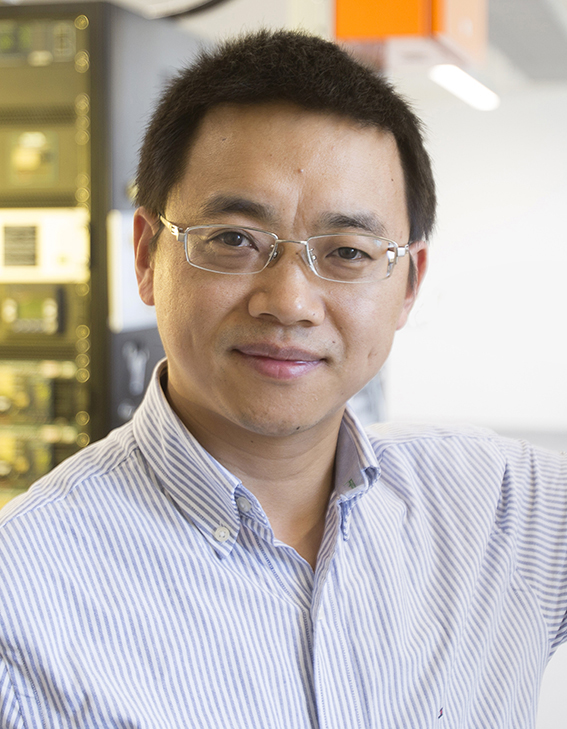}}]{Daoyi Dong}
	received the B.E. degree in automatic control and the Ph.D. degree in engineering from the University of Science and Technology of China, Hefei, China, in 2001 and 2006, respectively.
	
	He was an Alexander von Humboldt Fellow at AKS, University of Duisburg-Essen, Duisburg, Germany.
	He was with the Institute of Systems Science, Chinese Academy of Sciences, Beijing, China, and with Zhejiang University, Hangzhou, China. He had visiting positions at Princeton University, NJ, USA; RIKEN, Wako-Shi, Japan; and The University of Hong Kong, Hong Kong. 
	He is currently a Scientia Associate Professor at the University of New South Wales, Canberra, ACT, Australia. His research interests include quantum control and machine learning.
	
	Dr. Dong was awarded the ACA Temasek Young Educator Award by the Asian Control Association and was a recipient of the International Collaboration Award and the Australian Post-Doctoral Fellowship from the Australian Research Council, and a Humboldt Research Fellowship from the Alexander von Humboldt Foundation of Germany. He is a Member-at-Large, Board of Governors, and was the Associate Vice President for Conferences and Meetings, IEEE Systems, Man and Cybernetics Society. He served as an Associate Editor for the IEEE TRANSACTIONSON NEURAL NETWORKS AND LEARNING SYSTEMS from 2015 to 2021. He is currently an Associate Editor of the IEEE TRANSACTIONS ON CYBERNETICS and a Technical Editor of the IEEE/ASME TRANSACTIONS ON MECHATRONICS.
\end{IEEEbiography}

\end{document}